\documentclass[conference]{IEEEtran}
\IEEEoverridecommandlockouts
\usepackage{cite}
\usepackage{amsmath,amssymb,amsfonts}
\usepackage{algorithmic}
\usepackage{graphicx}
\usepackage{textcomp}
\usepackage{xcolor}
\usepackage{url}
\usepackage{booktabs}
\usepackage{enumitem}
\def\BibTeX{{\rm B\kern-.05em{\sc i\kern-.025em b}\kern-.08em
    T\kern-.1667em\lower.7ex\hbox{E}\kern-.125emX}}

\newcommand{\sds}[0]{\textsc{sds}}
\newcommand{\hri}[0]{\textsc{hri}}
    
\begin{document}

\title{Towards an Extensible Benchmark for \\ Spoken Dialogue with Social Robots}

\author{
\IEEEauthorblockN{Casey Kennington}
\IEEEauthorblockA{\textit{Department of Computer Science} \\
\textit{Boise State University}\\
Boise, Idaho, U.S.A. \\
caseykennington@boisestate.edu}
\and
\IEEEauthorblockN{Ross Mead}
\IEEEauthorblockA{\textit{Semio} \\
Los Angeles, \\ California, U.S.A. \\
ross@semio.ai}
\and
\IEEEauthorblockN{Saad Elbeleidy} 
\IEEEauthorblockA{\textit{Peerbots} \\
Arlington, Virginia, U.S.A. \\
saad@semio.community}
\and
\IEEEauthorblockN{Jesse Thomason}
\IEEEauthorblockA{\textit{School of Interactive Computing} \\
\textit{Georgia Institute of Technology}\\
Atlanta, Georgia, U.S.A. \\
jthomason7@gatech.edu}
}

\maketitle

\begin{abstract}
Language models provide a plug-and-play interface between humans and robots, but important challenges remain when speech, dialogue, fast interaction, and collaboration are required. We propose a benchmark for the community to use as a way to explore common spoken dialogue artifacts between robots and humans, including requests for clarification, interruptions, embodied signals (e.g., head nods or facial cues), and time constraints. We also explain our vision to extend the benchmark for other aspects of human-robot interaction that are important to the larger research community. To facilitate the benchmark, we further propose using \textit{Retico}, a real-time communication framework that fulfills important technical requirements to enable robots to have spoken dialogue capabilities.
\end{abstract}


\section{Introduction}
While research on large language models (LLMs) for human-robot interaction (\hri) has dramatically increased since the release of ChatGPT in 2022, it has become well-recognized in the broad research community that using LLMs for robots, particularly spoken dialogue with robots, has some major hurdles \cite{Marge2022-yf, Matuszek2025-yk}. Spoken dialogue systems (\sds) offer a better approach for humans to interact with robots than off-the-shelf LLMs because \sds\ research has shown effective ways to handle common-but-complex spoken interaction artifacts, such as misunderstandings, requests for clarification, turn-taking strategies, making decisions about what to say next, and interfacing with knowledge bases and external agents. Moreover, \sds\ research has informed advancements in models for multimodal interaction, speech recognition, language understanding, dialogue management, dialogue state-tracking, as well as natural language generation, each of which are important aspects of spoken dialogue communication. Given this background and the current state of research and technology, the SDS+HRI community has also highlighted a need for a common benchmark that can drive research towards a shared goal of natural communication between people and robots \cite{Kennington2024-ib}. 

\begin{table}[b]
    \centering
    \begin{tabular}{cc}
    \toprule
       \textbf{Robot-Ready SDS}  &  \textbf{LLM-Based Interaction} \\
       \midrule
speech-driven &text-driven \\
word-level granularity & sentence-level granularity\\
system requests clarification & system assumes understanding  \\
turn-taking is fluid & text-driven turns \\
live feedback & feedback not possible  \\
multimodal inputs are continuous & multimodal inputs are tokenized \\
\bottomrule
    \end{tabular}
    \vspace{6pt}
    \caption{Technical distinctions between robot-ready SDS and LLM-based interaction.}
    \label{tab:isdsvsllms}
\end{table}

What does it take to have natural dialogue with robots?  Past work identified five important technical requirements\cite{Kennington2020-qx}; specifically, a \textit{robot-ready \sds} must be: 
\begin{itemize}[nosep,noitemsep]
\item \textbf{modular}: robot components are modular and individual modules must be able to integrate with \sds\ modules;
\item \textbf{multimodal}: robots are \emph{situated} dialogue partners whose many sensors must be integrated with \sds\ speech input;
\item \textbf{distributive}: robot and \sds\ modules should easily communicate with each other in a distributed environment;
\item \textbf{incremental}: modules must be able to process input quickly and immediately; and
\item \textbf{aligned}: sensors must be synchronized (aligned) in time.
\end{itemize}

Table~\ref{tab:isdsvsllms} showcases some of the differences between robot-ready \sds\ requirements and standard LLM capabilities. Though ongoing work makes LLMs more capable of spoken, granular interaction, these differences are important as humans expect robots to be capable of more natural communication, particularly if the robot has anthropomorphic features \cite{Novikova2015-yx}. 

In building towards a community that can build and evaluate robot-ready \sds\ for use in HRI settings, we propose a benchmark and a framework, each detailed below. 

\section{Sketch of Potential Benchmark}
In this section, we explain one potential benchmark, beginning with desiderata, and then sketch an idea of what a potentially successful benchmark could look like. 

\subsection{Desiderata}
Members of the  SDS+HRI community identified several important desiderata for a proper benchmark for natural interaction with robots \cite{Kennington2024-ib}: 
\begin{itemize}[nosep,noitemsep]
    \item \textbf{multimodal}: the data are composed of speech, vision, robotic, and other modalities;
    \item \textbf{co-located}: the robot must have sensory access to the shared physical space;
    \item \textbf{high-stakes}: the user and robot must work together to complete a task that neither could do alone, where common dialogue artifacts like building common ground, providing explanations, clarification requests, and interruptions are common;
    \item \textbf{bi-directional}: interactions are not just commands from the user, and the robot also takes initiative when needed,
    \item \textbf{community-agnostic}: the benchmark should work on multiple robotic platforms; and
    \item \textbf{low-cost}: labs and research groups can easily and affordably acquire necessary items.
\end{itemize}

\subsection{Benchmark Outline}
Building on benchmarks that are meant for interactive dialogue with embodied agents \cite{Beyer2024-ri,Gao2023-ed,Shridhar2019-sz,Williams2020,StClair2011}, our proposed sketch of a benchmark is a collaborative game. The task is simplistic, but complicated enough to require important desired dialogue artifacts. Other games, tasks, or settings might be better suited for a benchmark, and we encourage the community to critique this benchmark and continue the discussion. 

\subsubsection{Base Task} The collaborative task will require a degree of asymmetrical knowledge and collaboration that neither the human nor the robot could do alone. 

\begin{figure}
    \centering
    \includegraphics[width=1.0\linewidth,angle=270]{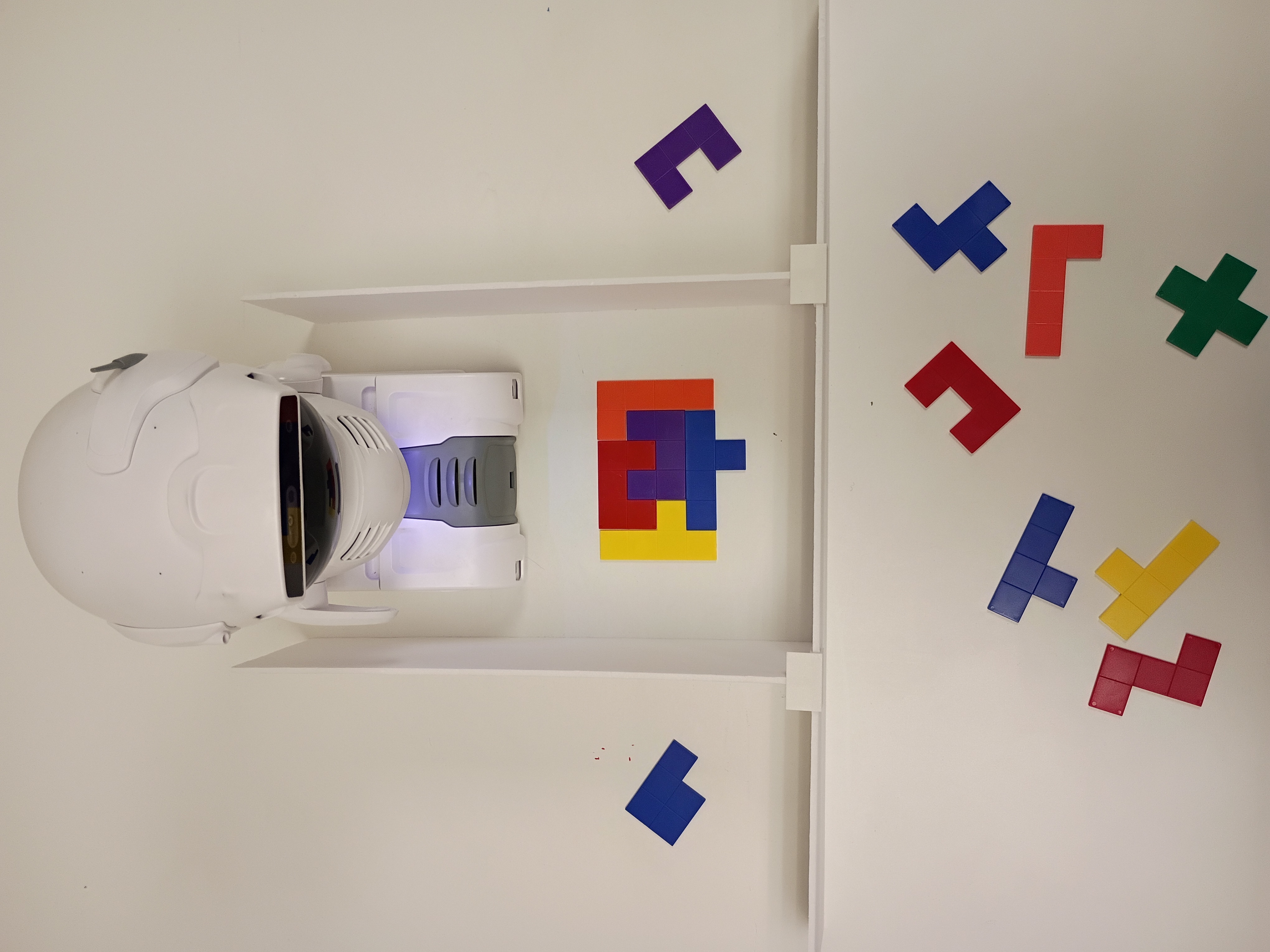}
    \caption{A top-down view of the base task. The game board is on the bottom which is visible to the instruction follower. Behind a divider is the target shape, the robot, and on either side of the robot are two puzzle pieces, at least one of them required for completing the task.}
    \label{fig:gameboard1}
\end{figure}

\begin{figure}
    \centering
    \includegraphics[width=1.0\linewidth,angle=270]{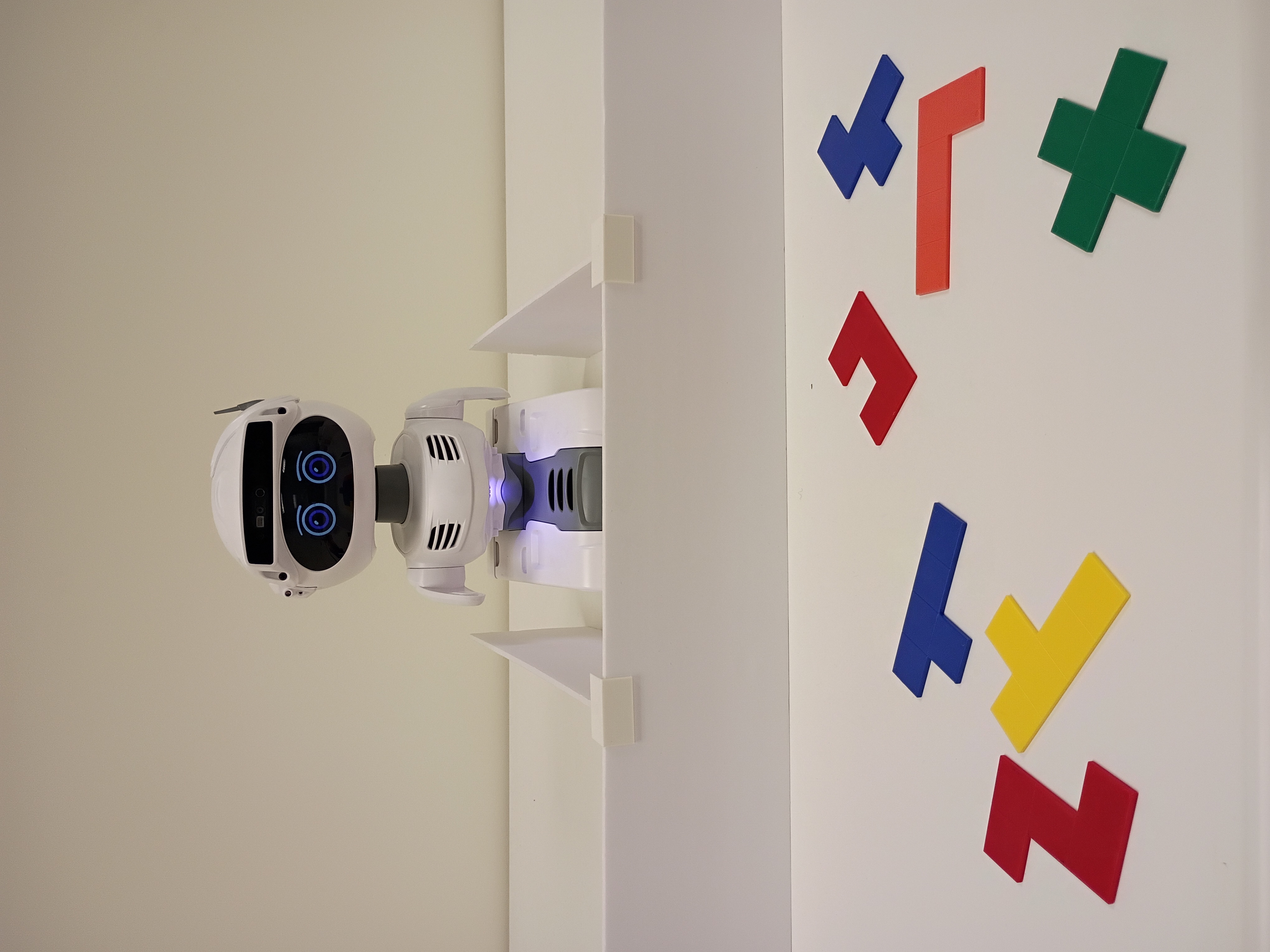}
    \caption{The human / instruction follower view of the game. The human can see the robot's face and arms, as well as the game board, but not the target shape or the two hidden pieces. The human can easily reach around the divider to replace one of the two hidden pieces.}
    \label{fig:gameboard2}
\end{figure}

Figures~\ref{fig:gameboard1} and \ref{fig:gameboard2} depict a possible setup using the Misty~II robot. There are two roles: the instruction \textit{giver} and the instruction \textit{follower} (though we stress that these roles serve only to help explain the benchmark; the task is meant to be more collaborative). In our explanation, the robot is the instruction \textit{giver} and the human is the \textit{follower}, but those roles can easily be switched. The goal is for the human and robot to collaborate to put the puzzle together to match the final image that the robot ``sees'' (the setup might require an extra camera that is not part of the robot). The human has access to some of the puzzle pieces on a game board with two dividers: one divider on the left, and one on the right, each hiding a separate puzzle piece from the player (the dividers can be anywhere as long as the participant cannot see the pieces, but they are within reach).  The robot is allowed to see everything including the final goal (top of the figure), a bird's eye view of the two puzzle pieces that the human cannot see behind the dividers (left and right in the figure), and the game board area where the human is working to construct the puzzle (center area of the figure).  The robot can explain to the human how to orient pieces (e.g., rotate left or right, flip, slide/shift left or right or up or down). The puzzle pieces themselves should have abstract shapes that require the human and the robot to initiate conversational grounding to determine how they describe or name the objects (e.g., pentomino puzzle pieces have shapes like letters L, T, or N). Furthermore, the puzzle pieces should have colors that might make some shapes less ambiguous, but not all puzzle pieces should have different colors. 

Not all puzzle pieces on the table will be used---there will be at least two distractor pieces. The robot also determines when the human should take the pieces on the left or the right dividers (if at all), with the rule that the human can only take one of those pieces if they replace it with another piece and they can only take those pieces once (a mistake results in an inability to complete the puzzle and a failed game). Players can organize the space to mark which objects are no longer eligible for placement in the boxes. By the end of the interaction, the puzzle should be constructed and the two distractor pieces should be out of view behind the two dividers. 

The robot minimally needs a camera that can move to look at the target, dividers, and game board with which the human is working. The robot can have any kind of face or use any kind of social capabilities, like eye gaze with the human, head nods, emotional displays, etc (e.g., \cite{Elbeleidy2025}). The robot can have hands for social interaction (e.g., deictic gestures~\cite{Williams2020,StClair2011}), but the goal is not to require the robot to perform complex physical manipulations.

\subsubsection{Metrics} Some proposed metrics include: (1)~number of puzzles started and completed, (2)~time to task completion, (3)~number of dialogue turns for both the human and the robot, (4)~number of initiations from human, (5)~number of initiations from the robot, and (6)~number of clarification requests from the human and robot, among others (e.g., interruptions, barge-ins). We recognize that some of these metrics will require dialogue act classification and other automated tracking of the dialogue, which the benchmark platform should provide.


\subsection{Extending the Base Task}

The benchmark is meant to be simple, yet complex enough to enable different research fields and specific research labs to test different aspects of human-robot spoken dialogue interaction. The benchmark can be extended to include additional constraints or settings that make it more difficult or require more collaboration. Additional metrics should be included for each extension as seen fit by the research community. For example, some research labs might not be concerned with dialogue aspects per~se and instead use a commercial language model as a placeholder, or they tune a local language model specifically for the task \cite{Williams2024-ng,Matuszek2025-yk}, but they wish to focus on gestures. Others might be directly concerned with acoustics, or dialogue aspects such as clarification, social attitudes, or robot manipulation. Researchers are able to hold some aspects of the system constant and range over the aspects of the interaction that interest them. We list some examples below. The broader research field of human-robot dialogue will have made marked progress if evaluated systems can accomplish the base task and all of the extensions. We welcome others in the community to suggest additional extensions. 

\subsubsection{Acoustics} Despite advances in speech recognition, there are still many challenges that need to be overcome for humans and robots to be able to speak to each other in settings outside of the lab. One challenge with human-robot dialogue, or any setting where a robot needs to listen to its environment is acoustics: microphone quality and placement, room layout including placement of the robot, ambient noise, decibel level of human speech, and other phenomena. An extension of the benchmark can be an \textit{acoustics} test suite that includes common acoustic challenges to the audio stream like adding the sound of white noise, robot actuators, alarms, a loud car driving by, decibel changes to the incoming speech, randomly changing words the speech recognizer outputs, etc. Researchers who would fit this benchmark extension include those working to address these challenges through improving speech recognition models, improving microphones and microphone arrays, experimenting with microphone placement, speaker diarization, robot voice filtering, etc.

\subsubsection{More Challenging Dialogue} To make the dialogue more challenging, one extension to the base task can be to introduce \textit{taboo} words that the robot is not allowed to say---for example, puzzle piece colors, common shape names, or directions (e.g., ``left''). This would require the robot to use other methods to explain itself. For example, if the robot wants to direct the human's attention to a specific piece that is red in color, but ``red'' is a taboo word, then the robot must find another way to communicate about the piece, either by using its shape, spatial placement, or by saying something in comparison, e.g., \textit{the piece that is the color of your shirt}. One of the metrics can be the number of taboo words the human is able to identify. The taboo words can be selected at the start of the session, or be dynamically introduced and revoked as the tasks progresses. 

\subsubsection{Robot Manipulation} The base task is meant to be a social task only, prioritizing spoken dialogue where the human completes all of the manipulation; however, we envision an extension to the base task that gives robotics researchers the opportunity to add robot movement and manipulation to the task. The first and most simple is that, when needed, the robot moves the hidden pieces from behind the divider to the human and takes the replacement piece. Moreover, research labs interested in robot locomotion can place the hidden pieces various locations in the room so the robot has to retrieve them. We also envision an extension where the robot and human can swap roles; the human becomes the instruction \textit{giver} who sees the target shape, the game board, and the hidden pieces and guides the robot (now the \textit{follower}), to construct the puzzle while the robot moves the pieces into place. A further extension might be that two robots take both roles and must communicate with each other to successfully put together the puzzle. A larger benchmark would require research lab teams to collaborate on pairing their robots and systems, with metrics extending to how the two robots work together.

\subsubsection{Varying Game Board} ``Pentomino objects'' are a good starting point due to their simple shapes and solid colors, but an extension of this could be more complex and commonplace objects identified by the community. For the base task, these can be small tabletop objects, such as a mug, a pen, a book, a bottle, a phone, and other everyday objects. The objects and size of the game board could go beyond tabletop and be much larger, such as cinder blocks or chairs, spanning an entire room or even multiple rooms. We also envision a scenario where the objects are more difficult to carry so the human and robot need to work together. The number of hidden and distractor objects could also vary, adding complexity. Finally, one extension could be that distractor objects are in another room that needs to be described by the robot to a human or to another robot that retrieves and replaces the object (but it is only allowed to enter that room once in a single game). 

\subsubsection{Gestures} The base task can be completed with dialogue alone, but humans often use methods other than complex descriptions to indicate objects that are in a shared space. An extension of the base task could therefore be to allow deictic and iconic gestures, with a potentially more complex game board starting state. Research labs who are interested in deictic gestures on robots would be able to test their theories using this extension, and labs interested in iconic gestures could use more complex puzzle pieces and configurations as needed.


\subsubsection{Player Attitudes} The base task assumes that the robot and human are collaborating with each other, but we also envision an extension that would enable additional player \textit{attitudes} that alter the social aspects of the interaction. For example, the human / instruction follower can be given instructions to be adversarial or reluctant to help (humans will very likely be commonly reluctant to interact with robots ``in the wild'' so it will be more ready to handle similar situations). If the robot / instruction giver detects reluctance or an adversarial attitude, it may need to switch to an alternative approach to social interaction---for example, to convince the human to help or to negotiate roles. Another attitude might include an adversarial instruction giver / robot that sometimes is reluctant to give information, or a robot that purposefully misguides or even lies about information (e.g., purposefully using the wrong word for a color feigning that it made a mistake) means the instruction follower has to ask clarification questions and adjust their strategy. Other attitudes include a human who is afraid of robots, different age categories, human disgust regarding certain robot morphologies, viewing robots as tools rather than social partners, over-trusting what the robot says even when it has made mistakes, feeling a sense of care over the robot (e.g., the robot mentions that it is depressed and needs help and doesn't want to play the game anymore). Most of these attitudes are on the side of the human, but many of these attitudes could also be modeled on the robot. 

\subsubsection{Discussion} The extensions described above are by no means exhaustive, but serve as a set of examples of how the benchmark can be extended to meet the needs of different communities. Taken together a robot that can complete all of the tasks would be a very capable robot, indeed. 

\section{Proposed Robot-Ready \sds\ Framework}
Michael~and~Möller~\cite{Michael2019-dx} introduced \textit{Retico} as a framework for real-time \sds\ conversations. \textit{Retico} is a realization of the \textit{incremental unit} theoretical model of interactive spoken dialogue \cite{Schlangen2011-wi}. It has twice been extended to include multimodal capabilities and bindings with robotic platforms \cite{Manaseryan2025-lb,Kennington2020-qx}. \textit{Retico}\footnote{\url{https://github.com/retico-team}} is free, open-source, and under active development with on-boarding videos and tutorials, as well as example systems and ongoing U.S. National Science Foundation support.
\textit{Retico} is as close to a robot-ready \sds\ framework as we have found, with ongoing additions to input sensors (e.g., microphones and cameras), object detection models, Hugging Face language model and agent integrations, benchmark bindings for ALFRED \cite{Shridhar2019-sz}, incremental (i.e., word-level) automatic speech recognizers, and robot bindings for the Cozmo, Vector, Misty II, and Booster K1 platforms, with more in development. Recent work has showcased the flexibility and usefulness of \textit{Retico} under different settings and research goals. \cite{Levandovsky2025-an,Manaseryan2026-sk}. 

\textit{Retico} is well-suited for the benchmark because it allows important dialogue capabilities, such as barge-ins, interruptions, and requests for clarification. Labs can use existing modules for speech recognition or robot support while maintaining flexibility to pursue their own research questions on different aspects of the system and robot. We assert that the benchmark platform should provide a baseline system with a dialogue task manager to control robot actions. 

\section{Conclusion}
We propose a benchmark for human-robot dialogue that fulfills important requirements for the community, yet allows flexibility across different research lab robot morphologies and capabilities. \textit{Retico} offers a way for robots to potentially interact more naturally with humans with incremental, distributed, modular, and multimodal capabilities. This benchmark is a starting point: we are planning workshops to take place in the near future to elicit feedback and input from the broader community and begin evaluations that use the benchmark.

\bibliographystyle{IEEEtran}
\bibliography{paperpile,custom}

\end{document}